\documentclass[utf8]{FrontiersinHarvard} % for articles in journals using the Harvard Referencing Style (Author-Date), for Frontiers Reference Styles by Journal: https://zendesk.frontiersin.org/hc/en-us/articles/360017860337-Frontiers-Reference-Styles-by-Journal
\usepackage{url,hyperref,microtype}
\usepackage[onehalfspacing]{setspace}

\usepackage{graphicx}

\usepackage{amsmath}
\usepackage{amsthm} 
\usepackage{amssymb}
\usepackage{bm}
\usepackage{bbm}
\usepackage{adjustbox}
\usepackage{subfig}
\newcommand*\rot{\rotatebox{90}}

\usepackage{algorithm}% http://ctan.org/pkg/algorithms
\usepackage{algpseudocode}%

\usepackage{graphicx}  % DO NOT CHANGE THIS
\usepackage{natbib}  % DO NOT CHANGE THIS AND DO NOT ADD ANY OPTIONS TO IT

\usepackage{placeins}
 
\usepackage{caption} % DO NOT CHANGE THIS AND DO NOT ADD ANY OPTIONS TO IT
\usepackage{algorithm}% http://ctan.org/pkg/algorithms
\usepackage{algpseudocode}% http://ctan.org/pkg/algorithmicx
\usepackage{tabularx}
\usepackage{verbatim}
\usepackage{bm}
\usepackage{lipsum}
\usepackage{tabularx}
\usepackage{mwe} %<- For dummy image
\usepackage[utf8]{inputenc} % allow utf-8 input
\usepackage{url}            % simple URL typesetting
\usepackage{booktabs}       % professional-quality tables
\usepackage{nicefrac}       % compact symbols for 1/2, etc.
\usepackage{microtype}      % microtypography
\usepackage{times}
\usepackage{epsfig}
\usepackage{graphicx}
\usepackage{amsmath}
\usepackage{amsthm}
\usepackage{mathrsfs} 
\usepackage{amssymb}
\usepackage{wrapfig}
\usepackage{lscape}
\usepackage{rotating}
\usepackage{subfig}
 
\usepackage{wrapfig}

\usepackage{bm}

\usepackage{setspace}
\usepackage{xspace}

\usepackage{breakcites}

\usepackage{xcolor}
\newtheorem{theorem}{Theorem}

\def\firstAuthorLast{Stan and Rostami} %use et al only if is more than 1 author
\def\Authors{Serban Stan and Mohammad Rostami}
\def\Address{University of Southern California, Los Angeles, CA, USA  }
\def\corrAuthor{Mohammad Rostami}

\def\corrEmail{rostamim@usc.edu}

\begin{document}
\onecolumn
\firstpage{1}

\title[OnlineUDA for Semantic Segmentation]{Online Continual Domain Adaptation for   Semantic Image Segmentation Using Internal Representations} 

\author[\firstAuthorLast ]{\Authors} %This field will be automatically populated
\address{} %This field will be automatically populated
\correspondance{} %This field will be automatically populated

\extraAuth{}% If there are more than 1 corresponding author, comment this line and uncomment the next one.
%\extraAuth{corresponding Author2 \\ Laboratory X2, Institute X2, Department X2, Organization X2, Street X2, City X2 , State XX2 (only USA, Canada and Australia), Zip Code2, X2 Country X2, email2@uni2.edu}

\maketitle

\begin{abstract}
 Semantic segmentation models trained on annotated data   fail to generalize well when the input data distribution changes over extended time period, leading to requiring re-training to maintain performance. Classic Unsupervised domain adaptation (UDA) attempts to address a similar  problem when there is target domain with no annotated data points through transferring knowledge from a source domain with annotated data.  We develop an online UDA algorithm for semantic segmentation of images that improves model generalization on unannotated domains in scenarios where source data access is restricted during adaptation. We perform model adaptation is  by minimizing the distributional distance  between the source latent features and the target features in a shared embedding space. Our solution promotes a shared domain-agnostic latent feature space between the two domains, which allows for classifier generalization on the target dataset. To alleviate the need of access to source samples during adaptation, we approximate the source latent feature distribution via an appropriate surrogate   distribution, in this case a Gassian mixture model (GMM). We evaluate our approach on well established semantic segmentation datasets and demonstrate it compares favorably against state-of-the-art (SOTA) UDA semantic segmentation methods.\footnote{Partial results  of this work were presented in the   AAAI Conference \cite{stan2021unsupervised}.}
\end{abstract}

\section{Introduction}

Recent progress in deep learning has led to developing  semantic segmentation algorithms that are being adopted in many real world tasks. Autonomous driving \cite{zhang2016instance, feng2020deep}, object tracking \cite{kalake2021analysis} or aerial scene parsing \cite{9568802} are just a few examples of these applications. Deep neural networks (DNNs) have proven indispensable for reaching above human performance in semantic segmentation tasks, given the ability of large networks to approximate complex decision functions \cite{DBLP:journals/corr/HeZRS15}. Training such networks however requires access to large continuously annotated datasets. Given that in semantic segmentation each image pixel requires a label, generating new labeled data for semantic segmentation tasks requires significant more overhead compared to regular classification problems. 

Domain Adaptation (DA) is a sub-field of AI which aims to allow model generalization for input distributions different from those observed in the training dataset \cite{wang2018deep}. Unsupervised Domain Adaptation (UDA) addresses this problem for instances where the deployment dataset lacks label information \cite{wilson2020survey}. This set of approaches is of especial interest for semantic segmentation tasks, where data annotation is expensive and time-consuming. In UDA, a model trained on a fully annotated \textit{source domain} needs to generalize on an unannotated \textit{target domain} with a different data distribution. A primary approach for achieving domain generalization is learning a shared embedding feature space between the source and target in which the domains become similar. If domain agnostic features are learnt, a semantic classifier trained on source data will maintain predictive power on target data. While this high level approach is shared among various UDA frameworks, different methods have been proposed for achieving this goal.

A common line of research achieves domain alignment by the use of adversarial learning \cite{goodfellow2020generative}. A feature extractor produces latent feature embeddings for the source and target domains, while a domain discriminator is tasked with differentiating the origin domain of the features. These two networks are trained adversarially, process which leads the feature extractor to learn a domain invariant feature representation upon training completion. There is a large body of UDA work following this methodology \cite{hoffman2018cycada, benjdira2019unsupervised, hoffman2016fcns, hong2018conditional, chen2018semantic,jian2023unsupervised}. A different set of approaches attempts direct distribution alignment between the source and target domains. Distribution alignment can then be achieved by minimizing an appropriate distributional distance metric ~\cite{wu2018dcan,zhang2017curriculum,gabourie2019learning,zhang2019category,lee2019sliced,yang2020fda,rostami2023domain,rostami2021cognitively}, such as $l2$-distance, $KL$-divergence or Wasserstein Distance.

While both types of approaches are able to obtain state of the art (SOTA) results on UDA semantic segmentation tasks, most methods assume simultaneous access to both source and target data samples. This benefits model stability during adaptation, as source domain access ensures a gradual shift of the decision function. However, in real world settings there are many situations where concomitant access to both domains cannot be achieved. For instance, datasets may need to be stored on different servers~\cite{stan2022secure} due to latency constraints \cite{9430906} or data privacy requirements \cite{LI2020101765,stan2021privacy}. To adhere to these settings, UDA has been extended to situations where the source domain data is no longer accessible during adaptation. This class of methods is named \textit{source-free adaptation}, and provides a balance between accuracy and privacy \cite{kundu2020universal, kim2021domain}. Compared to regular UDA, source-free UDA is less explored. Our approach addresses source-free adaptation, making it a suitable algorithm for scenarios where data privacy is an issue. Moreover, our proposed method is based on common DNN architectures for semantic segmentation, and requires little parameter fine-tuning compared to adversarial approaches. 

{\em \bf Contributions:} We propose a novel algorithm that performs source-free UDA for semantic segmentation tasks. Our approach eliminates the need for access to source data during the adaptation phase, by approximating the source domain via a internal intermediate distribution~\cite{rostami2021lifelong,rostami2020generative}. During adaptation, our method aligns the target and intermediate domain via a suitable distance metric to ensure classifier generalization on target features. We demonstrate the performance of our method on two benchmark semantic segmentation tasks, GTA5$\rightarrow$CITYSCAPES and SYNTHIA$\rightarrow$CITYSCAPES, where the source datasets are composed of computer generated images, and the target datasets are real world segmented images. We offer theoretical justification for our algorithms' performance, proving our approach minimizes target error under our adaptation framework. 

\section{Related Work}

We provide an overview of semantic segmentation algorithms, as well as describe recent UDA and source-free UDA approaches for this setting.

\subsection{Semantic Segmentation}

Compared to image classification problems, semantic segmentation tasks are more complicated because we require each pixel of an image to receive a label, which is part of a set of semantic categories. As each image dimension may have thousands of pixels, semantic segmentation models require powerful encoder/decoder architectures capable of synthesizing large amounts of image data and encode the spatial relationships between the pixels well. Recent SOTA results for supervised semantic segmentation have thus been obtained by the use deep neural networks (DNNs), and in particular convolutional neural networks (CNNs) \cite{lecun1995convolutional}, which are specialized for image segmentation. While different architecture variants exists ~\cite{long2015fully, wang2020axial, deeplabv3plus2018, tao2020hierarchical}, approaches often rely on embedding images into a latent feature space via a CNN feature extractor, followed by an up-sampling decoder which scales the latent space back to the original input space, where a classifier is trained to offer pixel-wise predictions. The idea is if the extracted features can reconstruct the input image with a relatively high accuracy, then they carry an information content similar to the input distribution, yet in a lower dimensional space. Skip connections between different levels of the network ~\cite{ronneberger2015unet, lin2017feature}, using dilated convolutions ~\cite{chen2017rethinking} or using transformer networks as feature extractors instead of CNNs ~\cite{strudel2021segmenter} have been shown to improve supervised baselines.

While improvements in supervised segmentation are mostly tied to architecture choice and parameter fine-tuning, model generalization suffers when changes in the input distribution are made. This phenomenon is commonly referred to as \textit{domain shift} \cite{sankaranarayanan2018learning}. Changes in the input distortion translate into shifted extracted features that do not match the internal distribution learned by the DNN~\cite{rostami2019complementary}. This issue is common in application domains where the same model needs to account for multi-modal data, and the training set lacks a certain mode, e.g. daylight and night-time images \cite{romera2019bridging}, clear weather and foggy weather \cite{sakaridis2018model}, medical data obtained from different imaging devices and scanners \cite{guan2021domain}. Such differences in input data distributions between source and target domains greatly impact the generalization power of learnt models. When domain shift is present, source-only training may be at least three-fold inferior compared to training the same model on the target dataset \cite{hoffman2016fcns, hoffman2018cycada, lee2019sliced}. While it is possible to retrain the model to account for distribution shifts, we will require to annotate data again which can be time-consuming and expensive in semantic segmentation tasks \cite{rostami2018crowdsourcing}. Because label information is expensive to obtain, adding a cost to using such techniques, especially on new datasets. Weakly-supervised approaches explore the possibility of having access to limited label information after domain shift to reduce the data annotation requirement \cite{Wei_2018_CVPR, hung2018adversarial, 8693661}. However, data annotation is still necessary.

On the other hand, due to the limited label availability for semantic segmentation tasks, the use of synthetic images and labels has become an attractive alternative for training segmentation models even if domain shift is not a primary concern. The idea is to prepare a synthetic source dataset which can be annotated automatically. Semantic labels are easy to generate for virtual images, and a model trained on such images could then be used on real world data as  a starting point. Overcoming domain shift becomes the primary bottleneck for successfully applying such models to new domains. 
 
 \subsection{Unsupervised Domain Adaptation}
 
 Unsupervised domain adaptation (UDA) addresses model generalization in scenarios where target data label information is unavailable but there a source domain with annotated data that shares the same labels with the target domain problem. UDA techniques primarily employ a shared feature embedding space between the source and target domain in which the distributions of both domains are aligned. A majority of these methods achieve this goal by either using domain discriminators based on adversarial learning or direct source-target feature alignment based on metric minimization.

 \subsubsection{Adversarial Adaptation for UDA}
 
  Techniques based on adversarial learning employ the idea of domain discriminator, used in GANs \cite{goodfellow2014generative}, to produce a shared source/target embedding space. A discriminator is tasked with differentiating whether two image encodings originate from the same domain, or one is from the source and one is from the target. A feature encoder aims to fool the discriminator, thus producing source/target latent features more and more similar in nature as training progressed. Over the course of training, this leads the feature extractor producing a shared embedding space for the source and target data.
  
  In the context of UDA for semantic segmentation, \citet{luc2016semantic} employ an image segmentation model and adversarially train a semantic map generator, which uses a label map discriminator to penalize the segmentation network for producing label maps more similar to the generated ones rather than the source ones. \citet{murez2017imagearxiv} use an intermediate feature distribution that attempts to capture domain agnostic features of both source and target datasets. To improve the domain agnostic representation, a discriminator is trained to differentiate whether an encoded image is part of the source or target domain. The encoder networks are then adversarially trained to fool the discriminator, resulting in similar embeddings between source and target samples. \citet{bousmalis2017unsupervised} develop a model for pixel level domain adaptation by creating a pseudo-dataset by making source samples appear as though they are extracted from the target domain. They use generative model to adapt source samples to the style of target samples, and a domain discriminator to differentiate between real and fake target data. \citet{hoffman2017cycada} employ the cycle consistency loss proposed in \cite{zhu2017unpaired} to improve the adversarial network adaptation performance. In addition to this, \cite{hoffman2017cycada} use GANs to stylistically transfer images between source and target domains, and use a consistency loss to ensure network predictions on the source image will be the same as in the stylistically shifted variant. \citet{saito2018maximum} use an approach based on a discriminator network without using GANs to attempt to mimic source or target data distributions. They propose the following adversarial learning process on a feature encoder network with two classification heads: (1) they first keep the feature encoder fixed and optimize the classifiers to have their outputs as different as possible, (2) they freeze the classifiers and optimize the feature encoder such that both classifiers will have close outputs. \citet{sankaranarayanan2018learning} employs an image translation network, that is tasked with translating input images into the target domain feature space. A discriminator is tasked with differentiating source images from target images passed through the network, and a similar procedure is done for target images. A pixel-level cross entropy loss ensures the network is able to perform semantic segmentation. \citet{lee2019sliced} use a similar idea to \cite{saito2018maximum} in that a network with two classifiers is used for adaptation. The feature extractor and classification heads are trained in an alternating fashion. The work of \cite{lee2019sliced} differentiates itself by employing an approximation of optimal transport to compute these discrepancy metrics, leading to improved performance over \cite{saito2018maximum}.
 
 \subsubsection{Adaptation by distribution alignment}
 
 Adaptation methods using direct distribution alignment share the same goal as adversarial methods. However, distribution alignment is reached by directly minimizing an appropriate distributional distance metric between the source and target embedding feature distributions. 
 
 \citet{wu2018dcan} propose an image translation network that takes as input source and target images, and outputs source images in the style of the target domain. Their proposed architecture does not use adversarial training, rather is based on the idea that in order for stylistic transfer to be achieved, domain mean and variance should be similar at different levels of abstraction throughout the translation network. They achieve this goal by minimizing $\ell_2$-distance in the feature space at various levels of abstraction. \citet{zhang2017curriculum} develop a method for semantic segmentation adaptation by observing that a source trained model should produce the same data statistics on the target domain as present in the annotated source distribution. Examples include label distribution or pixels of a certain class clustering around specific regions in an image. Pseudo-labeling is used to estimate these statistics \cite{wu2023unsupervised}. To enforce similarity in the output of a source-trained model to the estimated target statistics, KL-divergence is used as a minimization metric. \citet{zhang2019category} employ an adaptation framework based on the idea that source and target latent features should cluster together in similar ways. They use a pseudo-labeling approach to produce initial target labels, followed by minimizing the distance between class specific latent feature centroids between source and target domains. The minimization metric of choice is $\ell_2$-distance. For improved performance, they use category anchors to align the adaptation process. \citet{gabourie2019learning} propose an adaptation method based on a shared network encoder between source and target domains. Their model is trained by minimizing cross entropy loss for the source samples, and is tasked with minimizing the distance between source and target embeddings in the latent feature space. To achieve this goal, Sliced Wasserstein Distance is minimized between the source and target embeddings, leading to improved classifier performance on target samples. 
 
The expectation that continuous access to source data is guaranteed when performing UDA is not always true, especially in the case of privacy sensitive domains. This setting of UDA has been previously explored by methods that do not employ DNNs ~\cite{dredze2008online,jain2011online,wu2016online}, and has recently become the focus of DNN based algorithms for image classification tasks \cite{Kundu_2020_CVPR, kim2021domain, saltori2020sf, Yang_2021_ICCV}. Source-free semantic segmentation has been explored relatively less compared to joint UDA adaptation approach. \citet{Kundu_2021_ICCV} employ source-domain generalization and target pseudo-labeling in the adaptation method. \citet{liu2021source} rely on self supervision and patch level optimization for adaptation. \citet{10.1145/3474085.3475482} allow models trained on synthetic data to generalize on real data by a mixture of positive and negative class inference. 

Our adaptation approach shares the idea of direct distribution alignment. As described previously, several choices for latent feature alignment have been previously explored, such as $l2$-distance \cite{wu2018dcan}, KL-divergence \cite{zhang2017curriculum} or Wasserstein Distance (WD) \cite{gabourie2019learning, lee2019sliced}. WD has been proven to leverage the geometry of the embedding space better than other distance metrics \cite{tolstikhin2017wasserstein}. Empirically, the behavior of using the Wasserstein metric has been observed to benefit the robustness of training deep models, such as in the case of the Wasserstein GAN \cite{arjovsky2017wasserstein}, or by improving the relevance of discrepancy measures, as reported by \cite{lee2019sliced}. One of the limitations of using the WD is   the difficulty of optimizing this quantity, as computing the WD distance requires solving a linear program. Therefore, we employ an approximation of this metric, the Sliced Wasserstein Distance (SWD), which maintains the nice   metric properties of the WD while allowing for an end-to-end differentiation in the optimization process. 

We base our source-free UDA approach on estimating the latent source embeddings via an internal distribution ~\cite{rostami2019learning,rostami2022increasing}. This approximation relies on the concept that a supervised model trained on $K$ classes will produce a $K$ modal distribution in its latent space. This property of the internal distribution allows us to perform adaptation without direct access to source samples. The idea is to approximate the internal distribution and then sample from the $K$ modal  distributional approximation to use them as a surrogate for the source domain distribution. The distribution approximation  introduces a small number of parameters into our model. Once we produce a pseudo-dataset from sampling the internal distribution, we align the target feature encodings by minimizing the SWD between the two data distributions. Our theoretical bounds demonstrate our approach leads to minimizing an upperbound for the target domain error.

\section{Problem Formulation}

Let $\mathcal{P}_{\mathcal{S}}$ be the data distribution corresponding to a source domain, and $\mathcal{P}_{\mathcal{T}}$ be similarly the data distribution corresponding to a target domain, with $\mathcal{P}_{\mathcal{S}}$ being potentially different from $\mathcal{P}_{\mathcal{T}}$. We consider a set of multi-channel images $\bm{X_S}$ is randomly sampled from $\mathcal{P}_{\mathcal{S}}$ with corresponding pixel-wise semantic labels $\bm{Y_S}$. Let $\bm{X_T}$ be a set of images sampled from $\mathcal{P}_{\mathcal{T}}$, where we don't have access to the corresponding labels $\bm{Y_T}$. We consider that both $\bm{X_S}$ and $\bm{X_T}$ are represented as images with real pixel values in $\mathbb{R}^{W \times H \times C}$, where $W$ is the image width, $H$ is the image height and $C$ is the number of channels. The labels $\bm{Y_S}, \bm{Y_T}$  share the same input space of label maps in $\mathbb{R}^{W \times H}$ which makes the two domain relevant. 

Our goal is to learn the parameters $\theta$ of a   semantic segmentation model $\phi_\theta(\cdot): \mathbb{R}^{W \times H \times C} \rightarrow \mathbb{R}^{W \times H}$ capable of accurately predicting pixel-level labels for images sampled from the target distribution $\mathcal{P}_{\mathcal{T}}$. We can formulate this  problem as a supervised learning problem, where our goal is to minimize the target domain empirical risk, achieved by $\theta^* = \arg\min_{\theta}\{\mathbb{E}_{\bm{x}^{t}\sim P_{\mathcal{T}}(X^{t})}(\mathcal{L}(f_{\theta}(x^{t}),y^{t})\}$, where $x^t \in \bm{X_T}, y^t \in \bm{Y_T}$. The difficulty of the above optimization stems from the lack of access to the label set $\bm{Y_T}$. To overcome this challenge, we instead are provided access to the labeled source domain $(\bm{X_S}, \bm{Y_S})$, and then sequentially the target domain $\bm{X_T}$. Many UDA algorithms consider that both domains are accessible simultaneously but the source-free nature of our problem requires that once the target images $\bm{X_T}$ become available, access to source domain information becomes unavailable. This assumption is a practical assumption because domain shift is often a temporal problem that arises after the initial training phase.

To achieve training a generalizable model for the target domain, we need to first train a model on the provided source dataset and then adapted to generalize well on the target domain. Let $N$ be the size of the source dataset $\bm{X_S}$, and let $(x_i^s,y_i^s)$ be the image/label pairs from $\bm{X_S}, \bm{Y_S}$. Consider $K$ to be the number of semantic classes and $\mathbbm{1}_a(b)$ denote the indicator function determining whether $a$ and $b$ are equal. Then, we learn the parameters that minimize empirical risk on the source domain by optimizing the standard cross entropy loss on the labeled dataset:

\begin{equation}
    \begin{split}
    & \hat{\theta} =\arg\min_{\theta}\{\frac{1}{N} \sum_{i=1}^N\mathcal{L}_{ce}(\phi_{\theta}(x_i^s),y_i^s)\} 
    \\&\mathcal{L}_{ce}(p, y) =- \frac{1}{W H} \sum_{w=1}^{W} \sum_{h=1}^{H} \sum_{k=1}^K \mathbbm{1}_{y_{wh}}(k) \log(p_{wh}),
    \end{split}
    \label{eq:source-training}
\end{equation} 

The optimization setup in Eq. \ref{eq:source-training} ensures model generalization on inputs sampled from $\mathcal{P}_\mathcal{S}$. In cases where $\mathcal{P}_\mathcal{T}$ differs from $\mathcal{P}_\mathcal{S}$ the model will not generalize on the target domain, due to domain shift. To account for domain shift, we need to map data points from both domains into an invariant feature space between the two domains, without joint access to $\bm{X_S}$ and $\bm{X_T}$. To this end, let $f(\cdot),g(\cdot),h(\cdot)$ be three parameterized sub-networks such that $\phi = f \circ g \circ h$. In this composition, $f:\mathbb{R}^{W \times H \times C} \rightarrow \mathbb{R}^{L}$ is an encoder sub-network, $g:\mathbb{R}^{L} \rightarrow \mathbb{R}^{W \times H}$ is an up-scaling decoder, and $\mathbb{R}^{W \times H} \rightarrow \mathbb{R}^{W \times H}$ is a semantic classifier, where $L$ represents the dimension of the latent network embedding.  In order to create a shared source-target embedding space, our goal is that the network $(f \circ g)(\cdot)$ embeds source and target samples in a shared domain-agnostic embedding space. Under this condition, the classifier $h(\cdot)$ trained on source domain samples will be able to generalize on target inputs. 

We  can make the shared embedding space domain-agnostic by direct distribution alignment between the embeddings of the two domains at the decoder output. As previously explored in literature \cite{gabourie2019learning}, a suitable distributional distance metric $\mathcal{D}(\cdot,\cdot)$ can be minimized between the source and target domain data points at the network $(f \circ g)(\cdot)$ output. However, because the source domains are inaccessible during model adaptation, we cannot compute the distribution distance between the two domains. Hence, directly minimizing $\mathcal{D}(f \circ g(\bm{X_S}), f \circ g(\bm{X_T}))$ is not feasible. We need to develop a solution that relaxes the need for access to the source domain samples during adaptation for domain adaptation. Our core idea is to benefit from another distirbution that can be served as a surrogate for the source domain distirbution. We describe our source domain approximation approach and the choice for $\mathcal{D}(\cdot,\cdot)$ in the next section.

\section{Proposed Algorithm}

\begin{figure*}[!htb]
    \centering
     \includegraphics[width= \textwidth]{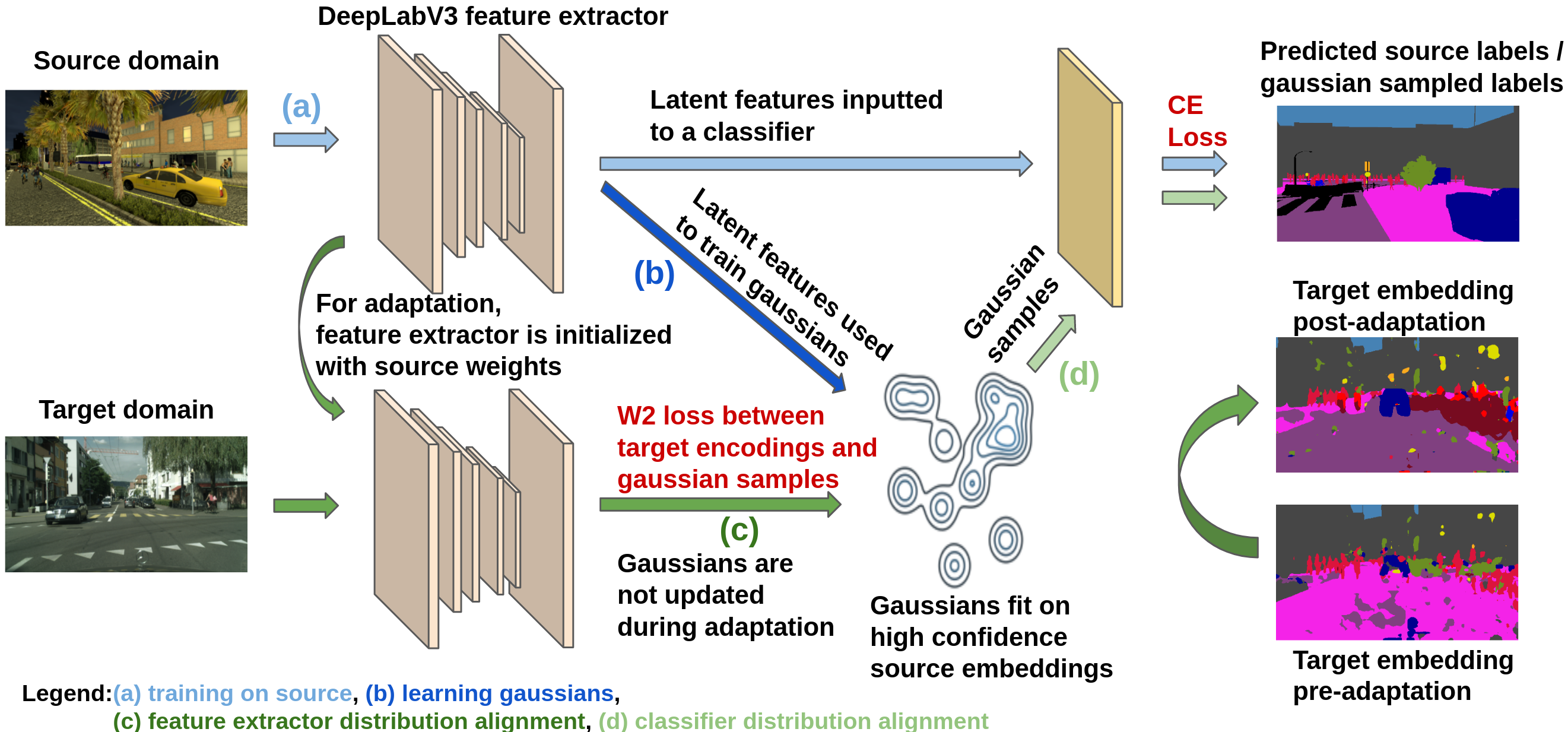}
    %  \caption{Block diagram of the proposed model adaptation approach: (a) initial model training using the source  domain labeled data (b) estimating the internal distribution as a GMM distribution in the embedding space (c) domain alignment is enforced by minimizing the distance between the internal distribution samples and the target unlabeled samples. }
    \caption{Diagram of the proposed model adaptation approach (best seen in color): (a) initial model training using the source  domain labeled data, (b) estimating the internal distribution as a GMM distribution in the embedding space, (c) domain alignment is enforced by minimizing the distance between the internal distribution samples and the target unlabeled samples, (d) domain adaptation is enforced for the classifier module to fit correspondingly to the GMM distribution.}
     \label{ISfig:blockdiagram}
\end{figure*}

We visually describe our method in Figure \ref{ISfig:blockdiagram}. The first step of our approach is to fully train a segmentation model on the labeled source domain. As training progresses on the source domain, the latent feature space will begin to cluster into $K$ clusters, where each of the clusters encode one of the semantic classes. If we use the output of the softmax layer as our embedding space,  the softmax classifier will be able to learn a linear decision function based on the decoder output which leads to high label accuracy at the end of this pre-training stage. After the source-training stage, we  approximate the source distribution via a learnt internal distribution. We use this as a surrogate for $f \circ g(\bm{X_S})$ during adaptation.

As linear separation in the latent space is reached, we can produce an approximation of the source domain distribution in the embedding space and thus relax the need for having access to the source samples during adaptation. We are interested in learning a $K$-modal approximation to the latent feature distribution at the decoder output, $f \circ g (\mathcal{P}_\mathcal{S})$. Let $p_k, 1 \leq k \leq K$ represent the component of $f \circ g (\mathcal{P}_\mathcal{S})$ corresponding to class $k$. For approximation purposes, we will   a Gaussian mixture model (GMM), with each semantic class approximated by  $T$ high  Gaussians components, i.e, the GMM would have $kT$ components in total. Our choice for this approximation method stems from the result by \cite{Goodfellow-et-al-2016} that concludes with sufficient Gaussians, any distribution can be approximated to vanishing error. While the GMM model is traditionally learned in an unsupervised fashion, we can leverage our knowledge of the source labels to partially supervise the process. As we have access to the source domain labeled data, we can directly identify which latent feature vectors correspond to each of the $K$ classes. Once the latent feature vectors are pooled for each class, a $T$ component GMM is learned using expectation maximization. During the learning process, we attempt to avoid inclusion of outlier elements in the GMMs, which may lead to decreased class separability in the latent space. This is a detrimental outcome for us, as an increased separability improves the performance of the classification layer. We thus only consider data samples which have high associated classifier confidence. Let $\tau$ be this confidence threshold, and let $S_{k} = \{ u_{i,j} | \exists (x^s,y^s), f \circ g (x^s)_{i,j} = u_{i,j}, y^s_{i,j}=k, f \circ g \circ h (x^s)_{i,j,k} > \tau \}$ be the set of source embedding feature vectors at the decoder output which have label $k$ and on which the classifier assigns to $k$ more than $\tau$ probability mass. Then, we use expectation maximization to learn $\alpha_{k,t}, \mu_{k,t}, \Sigma_{k,t}, 1 \leq t \leq T$ as the parameters of the $T$ components approximating class $k$. Thus, for each semantic class $k$, we model the latent feature distribution $p_k$ as:

\begin{equation}
    p_k(z) = \sum_{t=1}^T \alpha_{k,t} \mathcal{N} (z | \mu_{k,t}, \Sigma_{k,t})
    \label{eq:p_k}
\end{equation}

Learning a GMM approximation for each semantic class $k$ alleviates the need for source domain access during adaptation. Once adaptation stage starts, the source domain becomes unavailable, however we use the learnt GMM approximation distribution as a surrogate. We achieve classifier generalization on the target domain by minimizing a distributional distance metric between then GMM approximation and the target latent embeddings. For this purpose, consider the   dataset $\mathcal{Z} = (\bm{X_Z},\bm{Y_Z})$ produced by sampling from the GMM distribution, with $N^z = |\mathcal{Z}|$. Let $(x_i^z, y_i^z)$ be embedding/label pairs from this dataset. We achieve distribution alignment by empirically minimizing an appropriate distributional distance metric $D$ between samples from $\mathcal{Z}$ and from the target embeddings. In addition to distribution alignment, we need to account for shifts in the classifier input space between the internal distribution and the original source embedding distribution. We account for such shifts by fine tuning the classifier on labeled GMM samples. Our adaptation loss can be formalized as:

\begin{equation}
    \mathcal{L}_{adapt} = \mathcal{L}_{ce} (h(\bm{X_Z}), \bm{Y_Z}) + \lambda \mathcal{L}_{D} (f \circ g (\bm{X_T}), \bm{X_Z}) 
    \label{eq:adaptation-loss}
\end{equation}

for an appropriate choice of regularizer $\lambda$. 

The first loss term in Eq. \ref{eq:adaptation-loss} is the cross entropy classifier fine-tuning loss obtained for the GMM samples across the whole sampling dataset, i.e.,

\begin{equation}
    \begin{split}
        \mathcal{L}_{ce} (h(\bm{X_Z}), \bm{Y_Z}) = -&\frac{1}{N^z} \sum_{i=1}^{N^z} \frac{1}{W H} \sum_{w=1}^{W} \sum_{h=1}^{H} \sum_{k=1}^K \\ &\mathbbm{1}_{y_{i,wh}^z}(k) \log(h(x_{i}^z)_{wh})
    \end{split}
    \label{eq:ce-adaptation}
\end{equation}

where $(x_i^z, y_i^z)$ are the $i$'th GMM data point in the sampling dataset $\mathcal{Z}$. This loss terms helps maintaining model generalizability as we perform distribution alignment.

The second loss term in Eq. \ref{eq:adaptation-loss}, $\mathcal{L}_D$,  represents the distributional distance metric between the GMM in the latent space and the target domain data embedding vectors. We choose the Sliced Wasserstein Distance (SWD) as our choice for the distributional distance metric $D$. In the context of domain adaptation, several distribution alignment metrics have been previously used. \citet{wu2018dcan} propose an approach where the feature space between source and target images is made similar by directly minimizing the $\ell_2$-distance between feature vectors. $KL$-divergence has been used in domain adaptation \cite{yu2021divergence} to detect noisy samples or target samples from private classes. Wasserstein Distance (WD) has been explored as a distributional distance metric \cite{gabourie2019learning} by directly minimizing the metric on the output feature space for a source and target encoder. While study of appropriate distributional distance metrics is still ongoing, the WD aims to find the optimal way of moving mass between two distributions, and thus is tied to the geometry of the data. This has lead to the WD offering improved stability when used in a number of deep learning and domain adaptation tasks \cite{arjovsky2017wasserstein, kolouri2016sliced, solomon2015convolutional, bhushan2018deepjdot,rostami2019sar,rostami2019deep,xu2020reliable,li2020enhanced,rostami2020sequential}. The WD metric \cite{kolouri2019generalized} between two distributions $P$ and $Q$ are defined as:

\begin{equation}
    W_d(P,Q) = (\inf_{L \in \mathcal{L} (P,Q)} \int \lVert x - y \rVert^d dL(x,y))^{\frac{1}{d}}
    \label{eq:wasserstein-distance}
\end{equation}

where $\mathcal{L}(P,Q)$ represents all transportation plans between $P$ and $Q$, i.e., all joint distributions with marginals $P$ and $Q$. The $WD$ metric offers a closed form solution only  when $P$ and $Q$ are one dimensional distributions \cite{kolouri2019generalized}. For higher dimensions, we need to solve a linear program. While employing the WD has desirable properties, solving a linear program at every optimization step can lead to significant computational costs in the adaptation phase of UDA. To alleviate this issue we employ the SWD, an alternative for the WD that is fully differentiable, yet has a closed form formula. Computing the SWD between to high dimensional distributions involves repeatedly projecting them along random on dimensional projection directions, obtaining one dimensional marginals for which computation of the WD which has a closed form solution. This process  allows for an end-to-end differentiation via gradient based methods, such as Stochastic Gradient Descent \cite{bottou2018optimization}. Averaging one dimensional WD over sufficient random projction directions will produce a closed-form approximation to the high dimensional WD objective. We can the distributional distance term in Eq. \ref{eq:adaptation-loss} for two distributions $p,q$ as follows:

\begin{equation}
    \begin{split}
        \mathcal{L}_D(p,q) = \mathcal{SWD}_d(p,q) = \frac{1}{J} (\sum_{i=1}^J \lVert \gamma_i p - \gamma_i q \rVert^d)^\frac{1}{d}
    \end{split}
    \label{eq:adaptation-loss-swd}
\end{equation}

where $\mathcal{SWD}_d$ represents the $d$ order SWD, $J$ represents the number of random projection to be averaged, and $\gamma_i$ is one of the $J$ random projections. In our approach, we will choose $\mathcal{SWD}_2$ due to ease of computation and comparable performance to higher order choices of $d$. Pseudocode for our approach, named
Model Adaptation for Source-Free Semantic Segmentation
(MAS$^3$) is provided in Algorithm \ref{algorithm}.

\begin{algorithm}[t]
\caption{$\mathrm{MAS^3}\left (\lambda , \tau \right)$\label{algorithm}} 
 {
\begin{algorithmic}[1]
\State \textbf{Initial Training}: 
\State \hspace{2mm}\textbf{Input:} source domain dataset $\mathcal{D}_{\mathcal{S}}=(\bm{X}_{\mathcal{S}},  \bm{Y}_{\mathcal{S}})$,
\State \hspace{4mm}\textbf{Training on Source Domain:}
\State \hspace{4mm} $\hat{ \theta}_0=(\hat{\bm{w}}_0,\hat{\bm{v}}_0\hat{\bm{u}}_0) =\arg\min_{\theta}\sum_i \mathcal{L}(f_{\theta}(\bm{x}_i^s),\bm{y}_i^s)$
\State \hspace{2mm}  \textbf{internal Distribution Estimation:}
\State \hspace{4mm} Estimate the GMM parameters
\State \textbf{Model Adaptation}: 
\State \hspace{2mm} \textbf{Input:} target dataset $\mathcal{D}_{\mathcal{T}}=(\bm{X}_{\mathcal{T}})$
\State \hspace{2mm} \textbf{Pseudo-Dataset Generation:} 
\State \hspace{4mm} $\mathcal{D}_{\mathcal{P}}=(\textbf{Z}_{\mathcal{P}},\textbf{Y}_{\mathcal{P}})=$
\State \hspace{12mm} $([\bm{z}_1^p,\ldots,\bm{z}_N^p],[\bm{y}_1^p,\ldots,\bm{y}_N^p])$, where:
\State \hspace{16mm} $\bm{z}_i^p\sim \hat{p}_J(\bm{z}), 1\le i\le N_p$
\State \hspace{17mm}$\bm{y}_i^p=
\arg\max_j\{h_{\hat{\bm{w}}_0}(\bm{z}_i^p)\}$, $p_{ip}>\tau$
\For{$itr = 1,\ldots, ITR$ }
\State draw random batches from $\mathcal{D}_{\mathcal{T}}$ and $\mathcal{D}_{\mathcal{P}}$
\State Update the model by solving Eq.~\eqref{eq:adaptation-loss}
\EndFor
\end{algorithmic}}
\end{algorithm} 

\section{Theoretical Analysis}

We prove Algorithm \ref{algorithm} can lead to improving the model generalization on the target domain by minimizing an upperbound for the empirical risk of the model on the target domain. For such a result, we need to tie model generalization on the source domain to the distributional distance between the source and target domains. For this purpose, we use the framework developed by \citet{redko2017theoretical} designed for upper bounding target risk with respect to  the distance between the source and target domains in the classic joint UDA process. We rely on the following Theorem 2 from \citet{redko2017theoretical} in our approach:

\begin{theorem}(\citet{redko2017theoretical})
    For the variables defined under Theorem \ref{theorem1}, the following distribution alignment inequality loss holds:

    \begin{equation}
        \begin{split}
            \epsilon_T \leq &\epsilon_S + W(\hat\mu_S, \hat\mu_T)  + \sqrt{\big(2\log(\frac{1}{\xi})/\zeta\big)}\big(\sqrt{\frac{1}{N^s}}+\sqrt{\frac{1}{N^t}}\big) +  e_{\mathcal{C}}(h^*)    
        \end{split}
        \label{eq:theorem2}
    \end{equation}
    \label{theorem2}
\end{theorem}

The above relation characterises target error after source training, and does not consider our specific scenario of using an intermediate distribution. We adapt this bound for Algorithm \ref{algorithm} to derive the following theorem:

\begin{theorem}
    Consider the space of all possible hypotheses $\mathcal{H}$ applicable to the proposed segmentation task. Let $\epsilon_S(h), \epsilon_T(h)$ represent the expected source and target risk for hypothesis $h$, respectively. Let $\hat\mu_S, \hat\mu_Z, \hat\mu_T$ be the empirical mean of the embedding space for the source, intermediate and target datasets respectively. Let $W(\cdot,\cdot)$ represented the Wasserstein distance, and let $\xi, \zeta$ be appropriately defined constants. Consider $e_{\mathcal{C}}(h)$ to be the combined error of a hypothesis $h$ on both the source and target domains, i.e. $\epsilon_S(h) + \epsilon_T(h)$, and let $h^*$ be the minimizer for this function. Then, for a model $h$, the following results holds:

    \begin{equation}
        \begin{split}
        \epsilon_T(h) \leq &\epsilon_S(h) + W(\hat\mu_S, \hat\mu_Z) + W(\hat\mu_Z, \hat\mu_T) +  \sqrt{\big(2\log(\frac{1}{\xi})/\zeta\big)}\big(\sqrt{\frac{1}{N^s}}+\sqrt{\frac{1}{N^t}}\big) + e_{\mathcal{C}}(h^*)
        \end{split}
        \label{eq:theorem1}
    \end{equation}
    \label{theorem1}
\end{theorem}

\textbf{Proof:}  We expand the second term of Eq. \ref{eq:theorem2}. Given $W(\cdot, \cdot)$ is a convex optimization problem, we can use the triangle inequality as follows

\begin{equation}
    W(\hat\mu_S, \hat\mu_T) \leq W(\hat\mu_S, \hat\mu_Z) + W(\hat\mu_Z + \hat\mu_T)
    \label{eq:wasserstein-triangle}
\end{equation}

Combining Eq. \ref{eq:wasserstein-triangle} and Eq. \ref{eq:theorem2} leads to the result in Theorem \ref{theorem1}.

The above results provides a justification  Algorithm \ref{algorithm} is able to minimize the right hand side of the Eq. \ref{eq:theorem1}. The first term is minimized during the initial training phase on the source domain. Note that, as expected, the performance on the target domain cannot be better than the performance on the source domain. We conclude that the model we select, should be a good model to learn the source domain. The second term represents the WD distance between the source and sampling dataset. This distance will be small if the GMM approximation of the source domain will be successful. As we explained before, if select a large enough $T$, we can make this term negligible. The third term is the WD distance e between the sampling dataset and the target domain dataset. This term is directly minimized by the adaptation loss that we use to align the distribution. The term $1 - \tau$ is a constant directly dependent on the confidence threshold $\tau$, which we choose close to $1$. The fourth term is directly dependent on the dataset size, and becomes small when a large number of samples is present. Finally, the last term is a constant indicating the difficulty of the adaptation problem.

% Next, we analyze the last term of Eq. \ref{eq:theorem2}. We are interested in deriving a bound that incorporate the intermediate distribution. Let $\epsilon_Z(h)$ be the expected error of hypothesis $h$ on the intermediate dataset. 

% \begin{equation}
%     \begin{split}
%         e_C(h^*) = &\epsilon_S(h^*) + \epsilon_T(h^*) \\
%             = &\epsilon_S(h^*) + \epsilon_Z(h^*) + \epsilon_T(h^*) - \epsilon_Z(h^*) \\
%             \leq & \epsilon_S(h^*) + \epsilon_Z(h^*) + |\epsilon_T(h^*) - \epsilon_Z(h^*)|
%     \end{split}
% \end{equation}

\section{Experimental Validation}

We validate the proposed algorithm using common UDA benchmarks for semantic segmentation. Our implementation code is available as a supplement at \url{ https://github.com/serbanstan/mas3-continual}.

\subsection{Experimental Setup}

\subsubsection{Datasets}

We follow the UDA literature to evaluate our approach. We consider three common datasets used in semantic segmentation literature: GTA5~\cite{richter2016playing}, SYNTHIA~\cite{ros2016synthia} and Cityscapes~\cite{cordts2016cityscapes}. Both GTA5 and SYNTHIA are datasets consisting of artificially generated street images, with 24966 and 9400 instances respectively. Cityscapes is composed of real world images recorded in several European cities, consisting of 2957 training images and 500 test images. We can see that the diversity of sizes of these datasets which demonstrates the challenge of data annotation for semantic segmentation tasks. For all three datasets, images are processed and resized to a standard shape of $512 \times 1024$. 

Following the literature, we consider two adaptation tasks designed to evaluate model adaptation performance when the training set consists of artificial images, and the test set consists of real world images. For both SYNTHIA$\rightarrow$Cityscapes and GTA5$\rightarrow$Cityscapes, we evaluate performance under two scenarios, when $13$ or $19$ semantic classes are available. 

\subsubsection{Implementation and Training Details}

We use a DeepLabV3 architecture \cite{chen2017rethinking} with a VGG16 encoder \cite{simonyan2014very} for our CNN architecture. The decoder is followed by a $1\times1$ convolution softmax classifier. We choose a batch size of $4$ images for training, and use the Adam optimizer with learning rate of $1e-4$ for gradient propagation. For adaptation, we keep the same optimizer parameters as for training. We choose $100$ projections in our SWD computation, and set the regularization parameter $\lambda$ to $.5$. We perform training for $100k$ iterations on the source domain and then fpr adaptation we perform $50k$ iterations.

When approximating the GMM components, we chose the confidence parameter $\tau$ to be $0.95$. We observe higher values of $\tau$ to be correlated with increased performance, as expected from our theorem, and conclude that a $\tau$ setting above $0.9$ will lead to similar target performance. 

We run our experiments on a NVIDIA Titan XP GPU. Given that our method relies on distributional alignment, the label distribution between target batches may vary significantly between different batches. As the batch distribution approaches the target label distribution as the batch size increases, we use the oracle label distribution per batch when sampling from the GMM, which can be avoided if sufficient GPU memory becomes present. Experimental code is provided with the current submission.

\subsubsection{Baselines for Comparison}

% To the best of our knowledge, there is no prior source-free model adaptation algorithm for performance comparison. For this reason, we compare $MAS^3$ against UDA algorithms based on joint training due to proximity of these works to our learning setting.
Source-free model adaptation algorithms for semantic segmentation have been only recently explored. Thus, due most UDA algorithms being designed for joint training, besides source-free approaches we also include both pioneer and recent UDA image segmentation method to be representative of the literature. We have compared our performance against the adversarial learning-based UDA methods:  GIO-Ada~\cite{chen2018learning}, ADVENT~\cite{vu2018advent}, AdaSegNet~\cite{tsai2018learning}, TGCF-DA+SE \cite{choi2019selfensembling},  PCEDA~\cite{yang2020phase}, and CyCADA \cite{hoffman2017cycada}. We have also included  methods that are based on direct distributional matching which are more similar to $MAS^3$: FCNs in the Wild~\cite{hoffman2016fcns},  CDA~\cite{zhang2017curriculum}, DCAN~\cite{wu2018dcan},  SWD~\cite{lee2019sliced}, Cross-City~\cite{chen2017discrimination}. Source-free methods include GenAdapt \cite{Kundu_2021_ICCV} and SFDA \cite{liu2021source}.

\subsection{Comparison Results}

% \begin{table*}
%     \centering
%     % \subfloat[Office-home]{
%             % \adjustbox{valign=t, max width=\textwidth}{
%                 % \setlength\tabcolsep{2pt}
%                 \begin{tabular}{ |c|cccccccccccc|c| } 
%                     \hline
%                     \textbf{Method} & \textbf{A$\rightarrow$C} & \textbf{A$\rightarrow$P} & \textbf{A$\rightarrow$R} & \textbf{C$\rightarrow$A} & \textbf{C$\rightarrow$P} & \textbf{C$\rightarrow$R} & \textbf{P$\rightarrow$A} & \textbf{P$\rightarrow$C} & \textbf{P$\rightarrow$R} & \textbf{R$\rightarrow$A} & \textbf{R$\rightarrow$C} & \textbf{R$\rightarrow$P} & \textbf{Avg.} \\
%                     \hline
%                     Source only &49.4 &69.1 &74.8 &64.5 &63.5 &65.8 &51.4 &50.2 &73.4 &52.1 &44.1 &78.5 &61.4 \\
%                     MAS$^3$ (Ours) &55.9 &75.1 &79.3 &64.8 &72.2 &72.9 &54.2 &56.7 &79 &56.2 &51.9 &80.2 &66.5 \\
%                     \hline
%                 \end{tabular}
%             % }
%         % }
%     \caption{SUDA results on the Office-home dataset.}
%     \label{table:main-results-shared}
% \end{table*}

\subsection{SYNTHIA$\rightarrow$Cityscapes Task} 

We provide quantitative and qualitative results for this task in Table \ref{table:synthia}. We report the performance our method produces on the SYNTHIA$\rightarrow$CITYSCAPES adaptation task along with other baselines.
 Notably, even when confronted with a more challenging learning setting, $MAS^3$ demonstrates superior performance compared to the majority of classic UDA methods that have access to the source domain data during model adaptation. It is essential to highlight that some recently developed UDA methods leveraging adversarial learning surpass our approach in performance; however, it's worth noting that these methods often incorporate an additional form of regularization, aside from probability matching and are unable to address UDA when the source domain data is missing.

In an overarching evaluation, $MAS^3$ exhibits commendable performance, particularly when compared to UDA methods that rely on source samples. Furthermore, our method excels in specific crucial categories, such as the accurate detection of traffic lights, where it outperforms its counterparts. These results underscore the robustness and effectiveness of $MAS^3$ in handling challenging learning scenarios and achieving notable performance, especially in key object categories. We conclude that $MAS^3$ can be used to address classic UDA setting reasonably well.

\begin{table*}[ht!]
  \begin{adjustbox}{center}
  \small
  \scalebox{.8}{
      \begin{tabular} { @{} cr*{16}c   }
      \hline
       Method &Adv. &\rot{road} &\rot{sdwlk} &\rot{bldng} &\rot{light} &\rot{sign} &\rot{vgttn} &\rot{sky} &\rot{person} &\rot{rider}   &\rot{car} &\rot{bus} &\rot{mcycl} &\rot{bcycl} &mIoU \\
       \hline
        Source Only (VGG16)  &N &6.4 &17.7 &29.7 &0.0 &7.2 &30.3& 66.8& 51.1& 1.5& 47.3& 3.9& 0.1& 0.0 &20.2\\
        FCNs in the Wild &N &11.5 &19.6 &30.8 &0.1 &11.7 &42.3 &68.7 &51.2 &3.8 &54.0 &3.2 &0.2 &0.6 &22.9 \\
        CDA &N &65.2 &26.1 &74.9 &3.7 &3.0 &76.1 &70.6 &47.1 &8.2 &43.2 &20.7 &0.7 &13.1 &34.8 \\
        DCAN &N &9.9 &30.4 &70.8  &6.70 &23.0 &76.9 &73.9 &41.9 &16.7 &61.7  &11.5 &10.3 &38.6 &36.4 \\
        SWD &N &83.3 &35.4 &82.1 &12.2 &12.6 &83.8 &76.5 &47.4 &12.0 &71.5 &17.9 &1.6 &29.7 &43.5 \\
        
        Cross-City &Y &62.7 &25.6 &78.3 &1.2 &5.4 &81.3 &81.0 &37.4 &6.4 &63.5 &16.1 &1.2 &4.6 &35.7 \\
        GIO-Ada &Y &78.3 &29.2 &76.9 &10.8 &17.2 &81.7 &81.9 &45.8 &15.4 &68.0 &15.9 &7.5 &30.4 &43.0 \\
        ADVENT &Y &67.9 &29.4 &71.9 &0.6 &2.6 &74.9 &74.9 &35.4 &9.6 &67.8 &21.4 &4.1 &15.5 &36.6 \\
        AdaSegNet &Y &78.9 &29.2 &75.5 &0.1 &4.8 &72.6 &76.7 &43.4 &8.8 &71.1 &16.0 &3.6 &8.4 &37.6 \\
        TGCF-DA+SE &Y &90.1   &48.6 &80.7 &3.2 &14.3 &82.1 &78.4 &54.4 &16.4 &82.5 &12.3 &1.7 &21.8 &46.6 \\
        PCEDA &Y &79.7 &35.2 &78.7 &10.0 &28.9 &79.6 &81.2 &51.2 &25.1 &72.2 &24.1  &16.7 &50.4 &48.7 \\
        \hline
        SFDA &SF(Y) &81.9 &44.9 &81.7 &3.3 &10.7 &86.3 &89.4 &37.9 &13.4 &80.6 &25.6 &9.6 &31.3 &45.89 \\
        GenAdapt &SF(Y) &89.9 &48.8 &80.9 &19.5 &26.2 &83.7 &84.9 &57.4 &17.8 &75.6 &28.9 &4.3 &17.2 &48.9 \\
  \hline    
        MAS$^3$ (Ours)   &SF(N) &74.8 &51.6 &71.5 &20.4 &32.3 &73.0 &75.3 &48.9 &19.7 &66.3 &25.7 &10.1 &40.8 &47.0 \\
      \hline
      \end{tabular}
  }
  \end{adjustbox}
  \caption{Model adaptation comparison results  for the SYNTHIA$\rightarrow$Cityscapes task. We have used DeepLabV3~\cite{chen2017rethinking} as the feature extractor with a  VGG16~\cite{simonyan2014very} backbone. The first row presents the source-trained model performance prior to adaptation to demonstrate effect of initial knowledge transfer from the source domain.}
  \label{table:synthia}
\end{table*}

\subsection{GTA5$\rightarrow$Cityscapes Task} 
We present the quantitative outcomes for this particular task, detailed in Table  ~\ref{table:gta5}. It is noteworthy that we observe a more competitive performance in this task, and yet the overall trend in the performance comparison remains similar. These findings highlight the versatility of our proposed method, $MAS^3$. While our primary motivation lies in achieving source-free model adaptation, these results indicate that $MAS^3$ can effectively function as a joint-training UDA algorithm. We conclude our method manages to achieve state-of-the-art performance even in a setting involving a larger number of semantic classes. This capability underscores the robustness and adaptability of $MAS^3$ in diverse scenarios, making it a versatile solution that goes beyond its original focus on source-free model adaptation.

\begin{table*}[t]
  \begin{adjustbox}{center}
  \small
  \scalebox{.8}{
      \begin{tabular} { @{} cr*{20}c   }
      \hline
    %  \multicolumn{4}{|c|}{Country List} \\
    %  \hline
       Method &\rot{road} &\rot{sdwk} &\rot{bldng} &\rot{wall} &\rot{fence} &\rot{pole} &\rot{light} &\rot{sign} &\rot{vgttn} &\rot{trrn} &\rot{sky} &\rot{person} &\rot{rider} &\rot{car} &\rot{truck} &\rot{bus} &\rot{train} &\rot{mcycl} &\rot{bcycl} &mIoU \\
       \hline
        Source (VGG16) &25.9 &10.9 &50.5 &3.3 &12.2 &25.4 &28.6 &13.0 &78.3 &7.3 &63.9 &52.1 &7.9 &66.3 &5.2 &7.8 &0.9 &13.7 &0.7 &24.9 \\
        FCNs Wld. &70.4 &32.4 &62.1 &14.9 &5.4 &10.9 &14.2 &2.7 &79.2 &21.3 &64.6 &44.1 &4.2 &70.4 &8.0 &7.3 &0.0 &3.5 &0.0 &27.1 \\
        CDA &74.9 &22.0 &71.7 &6.0 &11.9 &8.4 &16.3 &11.1 &75.7 &13.3 &66.5 &38.0 &9.3 &55.2 &18.8 &18.9 &0.0 &16.8 &14.6 &28.9 \\
        DCAN &82.3 &26.7 &77.4 &23.7 &20.5 &20.4 &30.3 &15.9 &80.9 &25.4 &69.5 &52.6 &11.1 &79.6 &24.9 &21.2 &1.30 &17.0 &6.70 &36.2 \\
        SWD &91.0 &35.7 &78.0 &21.6 &21.7 &31.8 &30.2 &25.2 &80.2 &23.9 &74.1 &53.1 &15.8 &79.3 &22.1 &26.5 &1.5 &17.2 &30.4 &39.9 \\
        
        CyCADA &85.2 &37.2 &76.5 &21.8 &15.0 &23.8 &22.9 &21.5 &80.5 &31.3 &60.7 &50.5 &9.0 &76.9 &17.1 &28.2 &4.5 &9.8 &0.0 &35.4 \\
        ADVENT &86.9 &28.7 &78.7 &28.5 &25.2 &17.1 &20.3 &10.9 &80.0 &26.4 &70.2 &47.1 &8.4 &81.5 &26.0 &17.2 &18.9 &11.7 &1.6 &36.1 \\
        AdaSegNet &86.5 &36.0 &79.9 &23.4 &23.3 &23.9 &35.2 &14.8 &83.4 &33.3 &75.6 &58.5 &27.6 &73.7 &32.5 &35.4 &3.9 &30.1 &28.1 &42.4 \\
        TGCF-DA+SE &90.2 &51.5 &81.1 &15.0 &10.7 &37.5 &35.2 &28.9 &84.1   &32.7 &75.9 &62.7 &19.9 &82.6 &22.9 &28.3 &0.0 &23.0 &25.4 &42.5 \\
        PCEDA &90.2 &44.7 &82.0 &28.4 &28.4 &24.4 &33.7 &35.6 &83.7 &40.5 &75.1 &54.4 &28.2 &80.3 &23.8 &39.4 &0.0 &22.8 &30.8 &44.6 \\
        \hline
        SFDA &81.8 &35.4 &82.3 &21.6 &20.2 &25.3 &17.8 &4.7 &80.7 &24.6 &80.4 &50.5 &9.2 &78.4 &26.3 &19.8 &11.1 &6.7 &4.3 &35.86 \\
        GenAdapt &90.1 &44.2 &81.7 &31.6 &19.2 &27.5 &29.6 &26.4 &81.3 &34.7 &82.6 &52.5 &24.9 &83.2 &25.3 &41.9 &8.6 &15.7 &32.2 &43.4 \\
        \hline
        MAS$^3$ (Ours)   &75.8 &55.6 &72.9 &20.9 &24.7 &20.5 &30.5 &39.8 &80.0 &36.9 &77.9 &51.9 &22.4 &77.3 &26.5 &45.2 &22.6 &18.8 &51.7 &44.8\\
      \hline
      \end{tabular}
  }
  \end{adjustbox}
  \caption{Domain adaptation results for different methods for the GTA5$\rightarrow$Cityscapes task.}
  \label{table:gta5}
\end{table*}

\subsection{Analytic Experiment} 

\begin{figure*}[ht]
    \centering
    \includegraphics[width=1\textwidth]{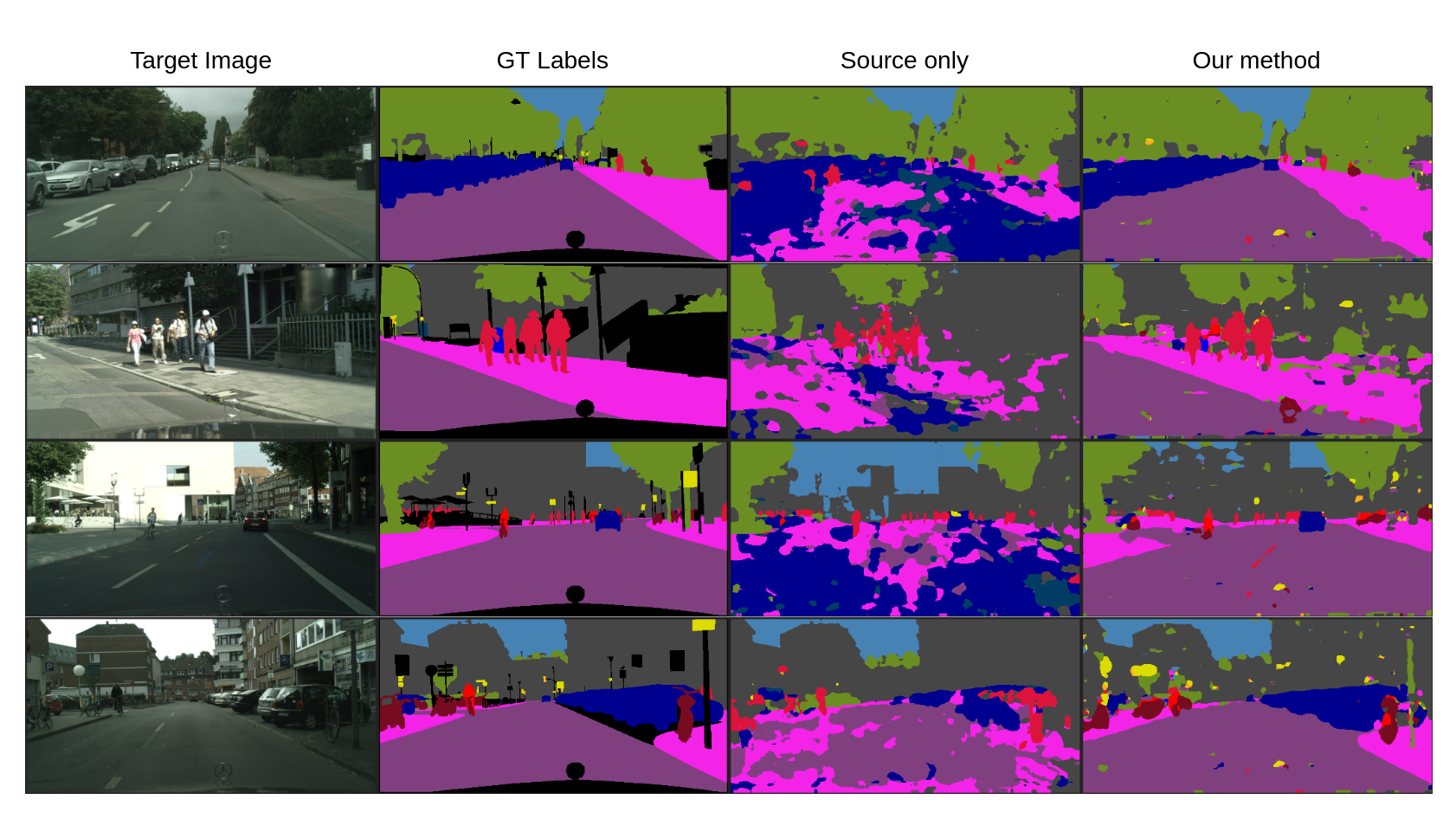}
    \caption{Qualitative performance: examples of the   segmented frames for SYNTHIA$\rightarrow$Cityscapes using the MAS$^3$ method. Left to right: real images, manually annotated images, source-trained model predictions, predictions based on our method.  }
    \label{figure:da}
\end{figure*}

\begin{figure*}[ht]
    \centering
    \includegraphics[width=.92\textwidth]{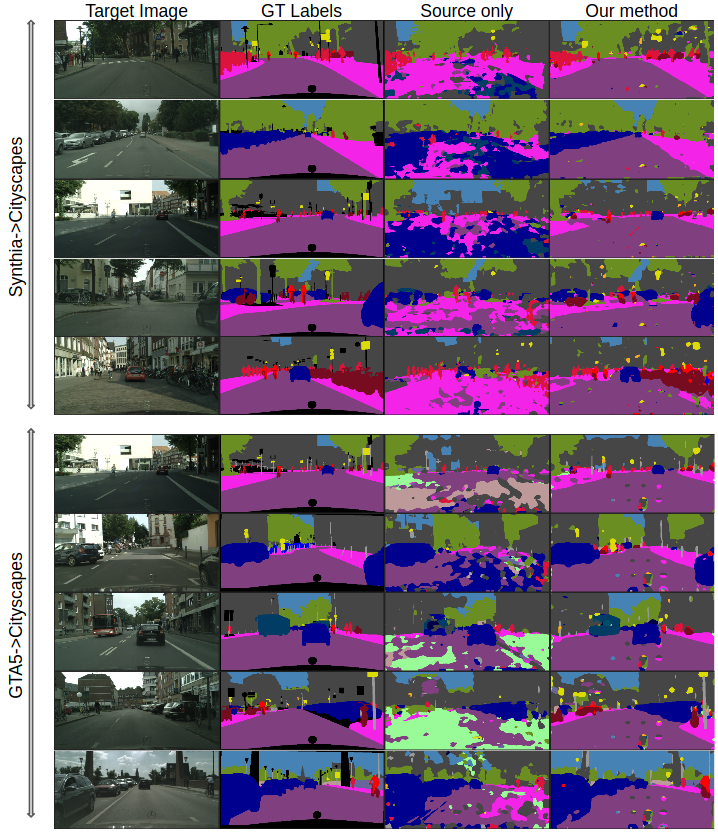}
    \caption{Qualitative performance: examples of the segmented frames for SYNTHIA$\rightarrow$Cityscapes and GTA5$\rightarrow$Cityscapes using the MAS$^3$ method. From left to right column: real images, manually annotated images, source-trained model predictions, predictions based on our method.  }
    \label{figure:da-extended}
\end{figure*}

We offer additional experiments to offer a better insight about our algorithm.
In Figure~\ref{figure:da}, we have provided visualizations of representative frames from the Cityscapes dataset for the SYNTHIA$\rightarrow$Cityscapes task. These frames are segmented using our model both prior to and after adaptation, and are juxtaposed with the corresponding ground-truth  manual annotations for each image. Through visual inspection, it becomes evident that our method brings about significant improvements in image segmentation, transitioning from source-only segmentation to post-adaptation segmentation. This improvement is particularly notable in semantic classes such as sidewalk, road, and cars for the model initially trained on GTA5 which are particularly important classes in autonomous driving applications.
The visual comparison highlights the considerable enhancement achieved in performance.
To further complement these findings, examples of segmented frames for the SYNTHIA$\rightarrow$Cityscapes task are included in Figure \ref{figure:da-extended}, revealing similar observations. These visualizations collectively underscore the effectiveness of our method in enhancing image segmentation across diverse datasets.

\begin{figure*}[t]
  \centering
    \begin{minipage}{0.32\textwidth}\includegraphics[width=\textwidth]{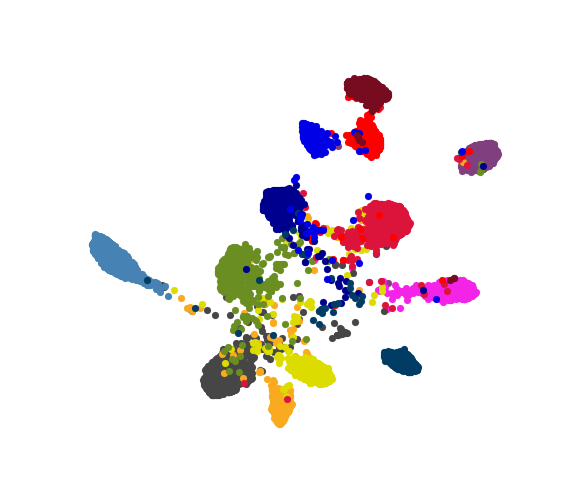}
        \\ \centering (a) GMM samples 
       % \label{figure:latent-features1}
    \end{minipage}
  \centering
    \begin{minipage}{0.32\textwidth}\includegraphics[width=\textwidth]{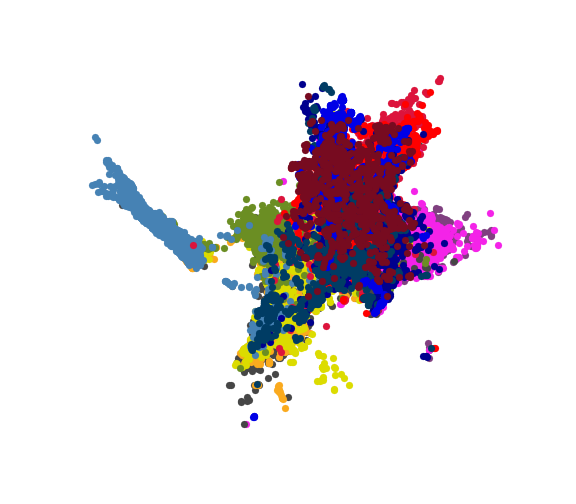}
        \\ \centering (b) Pre-adaptation 
       %\label{figure:latent-features2}
    \end{minipage}
       \begin{minipage}{0.32\textwidth}\includegraphics[width=\textwidth]{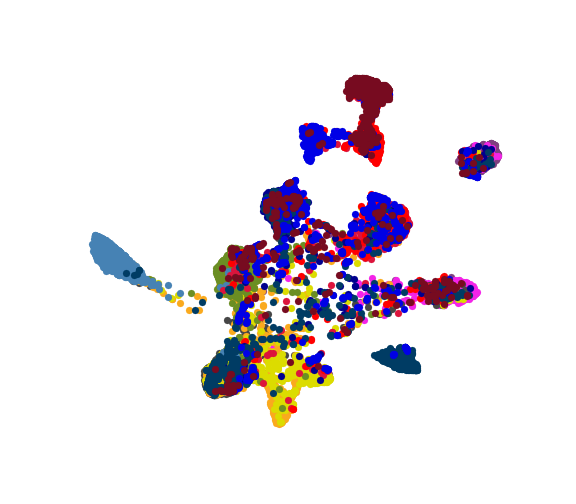}
           \centering
      \\ \centering (c) Post-adaptation 
       %  \label{figure:latent-features3}
    \end{minipage}
     \caption{Indirect distribution matching in the embedding space: (a) drawn samples  from the GMM trained on the SYNTHIA distribution, (b) representations of the Cityscapes validation samples prior to model adaptation (c) representation of the Cityscapes validation samples after domain alignment. }\label{figure:latent-features}
\end{figure*}

We study the effect our algorithm on data distirbution in the embedding space.
To validate the alignment achieved by our solution, we employed the UMAP~\cite{mcinnes2018umap} visualization tool to reduce the dimensionality of data representations in the embedding space to two for 2D visualization. Figure~\ref{figure:latent-features} visually represents samples from the internal distribution, along with the target domain data both before and after adaptation for the GTA5$\rightarrow$Cityscapes task. In this figure, each point corresponds to a single data point, and each color represents a semantic class cluster.
Upon comparing Figure~\ref{figure:latent-features} (b) and Figure~\ref{figure:latent-features} (c) with Figure~\ref{figure:latent-features} (a), a noticeable observation emerges. The semantic classes in the target domain exhibit greater separation and similarity to the internal distribution after the model adaptation process. This signifies a substantial reduction in domain discrepancy facilitated by $MAS^3$, where the source and target domain distributions align indirectly through the intermediate internal distribution in the embedding space, as originally anticipated.

% \begin{figure}[t!]
% \centering
%     \subfloat[GMM samples]{
%         % \includegraphics[width=.14\textwidth]{Figures/gaussian_embedding.png}
%         \includegraphics[width=.3\textwidth]{Figures/gaussian_embedding_v2.png}
%         \label{figure:latent-features1}
%     }
%     \subfloat[Pre-adaptation]{
%         % \includegraphics[width=.14\textwidth]{Figures/target_non_adapted.png}
%         \includegraphics[width=.3\textwidth]{Figures/target_non_adapted_v2.png}
%         \label{figure:latent-features2}
%     }
%     \subfloat[Post-adaptation]{
%         % \includegraphics[width=.14\textwidth]{Figures/target_adapted.png}
%         \includegraphics[width=.3\textwidth]{Figures/target_adapted_v2.png}
%         \label{figure:latent-features3}
%     }
%     \caption{Indirect distribution matching in the embedding space: (a) drawn samples  from the GMM trained on the SYNTHIA distribution, (b) representations of the Cityscapes validation samples prior to model adaptation (c) representation of the Cityscapes validation samples after domain alignment.}
%     \label{figure:latent-features}
% \end{figure}

\subsection{Sensitivity Analysis Experiments}

\begin{figure*}[t]
  \centering
    \begin{minipage}{0.32\textwidth}\includegraphics[width=\textwidth]{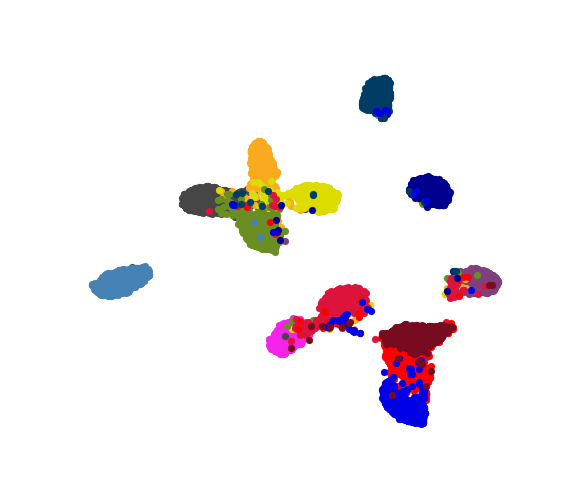}
        \\ \centering (a) $\tau=0$, mIoU=$41.8$  
     %   \label{figure:latent-features1}
    \end{minipage}
  \centering
    \begin{minipage}{0.32\textwidth}\includegraphics[width=\textwidth]{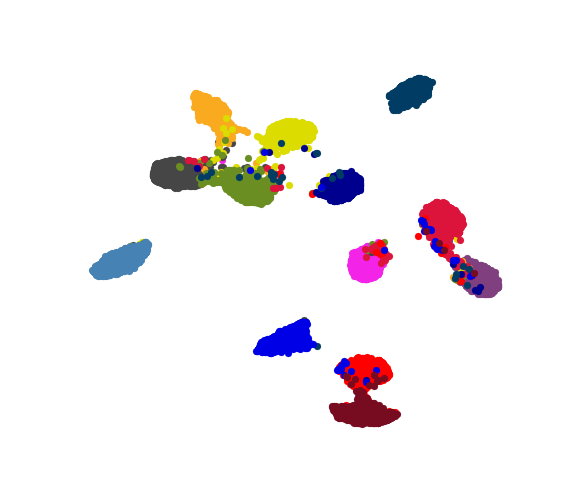}
       \\ \centering (b) $\tau=0.8$, mIoU=$42.7$  
   %    \label{figure:latent-features2}
    \end{minipage}
       \begin{minipage}{0.32\textwidth}\includegraphics[width=\textwidth]{Figures/target_adapted_v2.png}
           \centering
      \\ \centering (c) P$\tau=0.97$, mIoU=$44.7$ 
      %   \label{Figures/gaussians_rho=.97.png}
    \end{minipage}
     \caption{Ablation experiment to study effect of $\tau$ on the GMM learnt in the embedding space: (a) all samples are used; adaptation mIoU=41.8, (b) a portion of samples is used; adaptation mIoU=42.7, (c) samples with high model-confidence are used; adaptation mIoU=44.7 }\label{figure:rho-impact}
\end{figure*}

An inherent advantage of our algorithm, in contrast to methods relying on adversarial learning, lies in its simplicity and dependence on only a few hyperparameters. We study the sensitivity of performance with respect to these hyperparameters. The primary hyperparameters specific to our algorithm are  $\lambda$ and $\tau$ constants. Through experimentation, we have observed that the performance of $MAS^3$ remains stable with respect to the trade-off parameter $\lambda$. This stability is expected as the $\mathcal{L}_{ce}$ loss term  remains relatively small from the outset due to prior training on the source domain and then optimization mostly reduces the cross-domain alignment loss term.
We further delved into the impact of the confidence hyperparameter $\tau$. Figure~\ref{figure:rho-impact} visually illustrates the fitted Gaussian Mixture Model (GMM) on the source internal distribution for three different values of $\tau$. Notably, when $\tau=0$, the fitted GMM clusters appear cluttered. However, as we increment the threshold $\tau$ and selectively use samples for which the classifier demonstrates confidence, the fitted GMM represents well-separated semantic classes. This increase in interclass clusters in knowledge transfer from the source domain is evident as semantic classes become more distinctly defined.
This empirical exploration aligns with our earlier deduction regarding the significance of $\tau$, as outlined in our Theorem, thereby validating the theoretical analysis. The experimental findings underscore the robustness and effectiveness of our method across different hyperparameter configurations.

% \begin{figure}[t]
%     \centering
%     \subfloat[$\tau=0$, mIoU=$41.8$]{
%         \includegraphics[width=.3\textwidth]{Figures/gaussians_rho=0.png}
%     }
%    \subfloat[$\tau=0.8$, mIoU=$42.7$]{
%         \includegraphics[width=.3\textwidth]{Figures/gaussians_rho=.8.png}
%     }
%     \subfloat[$\tau=0.97$, mIoU=$44.7$]{
%         \includegraphics[width=.3\textwidth]{Figures/gaussians_rho=.97.png}
%     }
%     \caption{Ablation experiment to study effect of $\tau$ on the GMM learnt in the embedding space: (a) all samples are used; adaptation mIoU=41.8, (b) a portion of samples is used; adaptation mIoU=42.7, (c) samples with high model-confidence are used; adaptation mIoU=44.7}
%     \label{figure:rho-impact}
% \end{figure}

\begin{figure*}[t]
    \centering
    \includegraphics[width=.4\textwidth]{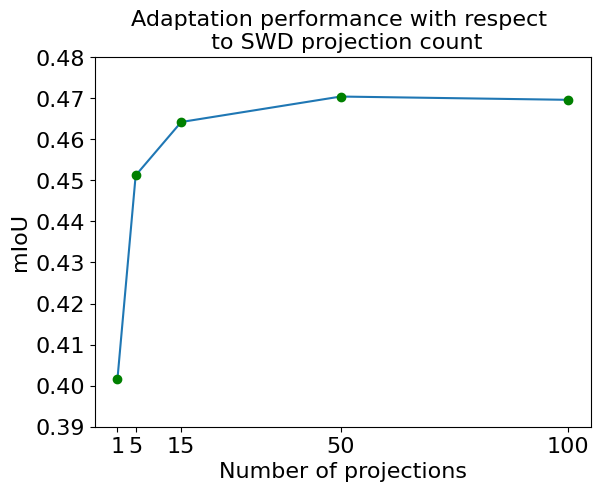}
    \caption{Performance accuracy versus the number of projections   used to compute   SWD using the SYNTHIA$\rightarrow$Cityscapes task.  }
    \label{figure:projection-performance}
\end{figure*}

We also extend our empirical investigation to analyze the balance between the quantity of projections employed for computing SWD for distribution alignment and the resulting UDA performance on the target domain. Results for this experiment is presented in Figure ~\ref{figure:projection-performance}. We observe that the performance remains   robust even with a modest number of projections, such as $5$. Moreover, empirical evidence indicates that the performance tends to plateau after approximately $50$ projections. It is noteworthy that the runtime complexity scales linearly concerning the number of projections, as the only operation involving the one-dimensional Wasserstein computations is the subsequent averaging process.
In consideration of these findings, the results presented throughout the rest of the paper are based on utilizing $100$ projections. This choice is made to ensure that we operate within a favorable regime concerning adaptation performance while maintaining a balanced runtime to compute SWD. By doing so, we aim to   achieve desirable performance with a decent computational load.

\section{Conclusions}

We   devised an algorithm tailored for adapting an image segmentation model to achieve generalization across new domains, a process facilitated solely through the use of unlabeled data for the target domain during training. At the core of our approach is the utilization of an intermediate multi-modal internal distribution, strategically employed to minimize the distributional cross-domain discrepancy within a shared embedding space. To estimate this internal distribution, we employ a parametric Gaussian Mixture Model (GMM) distribution. Through rigorous experimentation on benchmark tasks, our algorithm has demonstrated its effectiveness, yielding competitive performance that stands out even when compared to existing UDA algorithms rooted in joint-domain model training strategies. The results underscore the robustness and efficacy of our approach in achieving domain adaptation for image segmentation tasks, particularly in scenarios where only unlabeled data is available for training. Future exploration inlcudes partial domain adaptation settings in which the source and the target domain do not share the same classes.

\bibliographystyle{Frontiers-Harvard}
\bibliography{ref}

\end{document}